\documentclass[11pt, twocolumn]{article}

\usepackage[authoryear, sort&compress, round]{natbib}

\usepackage{geometry}

\usepackage{titlesec}
\titleformat{\section}{\large\bfseries}{\thesection}{1em}{}

\title{Talking About Large Language Models}

\author{
Murray Shanahan\\
Imperial College London\\
m.shanahan@imperial.ac.uk
}

\date{
December 2022\\
Revised February 2023
}

\begin{document}

\maketitle

\begin{abstract}
Thanks to rapid progress in artificial intelligence, we have entered an era when technology and philosophy intersect in interesting ways. Sitting squarely at the centre of this intersection are large language models (LLMs). The more adept LLMs become at mimicking human language, the more vulnerable we become to anthropomorphism, to seeing the systems in which they are embedded as more human-like than they really are. This trend is amplified by the natural tendency to use philosophically loaded terms, such as ``knows'', ``believes'', and ``thinks'', when describing these systems. To mitigate this trend, this paper advocates the practice of repeatedly stepping back to remind ourselves of how LLMs, and the systems of which they form a part, actually work. The hope is that increased scientific precision will encourage more philosophical nuance in the discourse around artificial intelligence, both within the field and in the public sphere.
\end{abstract}

\section{Introduction}

The advent of large language models (LLMs) such as Bert \citep{devlin2018bert} and GPT-2 \citep{radford2019language} was a game-changer for artificial intelligence. Based on transformer architectures \citep{vaswani2017attention}, comprising hundreds of billions of parameters, and trained on hundreds of terabytes of textual data, their contemporary successors such as GPT-3 \citep{brown2020language}, Gopher \citep{rae2021scaling}, and PaLM \citep{chowdhery2022palm} have given new meaning to the phrase ``unreasonable effectiveness of data'' \citep{halevy2009unreasonable}.

The effectiveness of these models is ``unreasonable'' (or, with the benefit of hindsight, somewhat surprising) in three inter-related ways. First, the performance of LLMs on benchmarks {\em scales} with the size of the training set (and, to a lesser degree with model size). Second, there are {\em qualitative leaps} in capability as the models scale. Third, a great many tasks that demand intelligence in humans can be {\em reduced to next token prediction} with a sufficiently performant model. It is the last of these three surprises that is the focus of the present paper.

As we build systems whose capabilities more and more resemble those of humans, despite the fact that those systems work in ways that are fundamentally different from the way humans work, it becomes increasingly tempting to anthropomorphise them. Humans have evolved to co-exist over many millions of years, and human culture has evolved over thousands of years to facilitate this co-existence, which ensures a degree of mutual understanding. But it is a serious mistake to unreflectingly apply to AI systems the same intuitions that we deploy in our dealings with each other, especially when those systems are so profoundly different from humans in their underlying operation.

The AI systems we are building today have considerable utility and enormous commercial potential, which imposes on us a great responsibility. To ensure that we can make informed decisions about the trustworthiness and safety of the AI systems we deploy, it is advisable to keep to the fore the way those systems actually work, and thereby to avoid imputing to them capacities they lack, while making the best use of the remarkable capabilities they genuinely possess.

\section{What LLMs Really Do}

As Wittgenstein reminds us, human language use is an aspect of human collective behaviour, and it only makes sense in the wider context of the human social activity of which it forms a part \citep{wittgenstien1953philosophical}.	A human infant is born into a community of language users with which it shares a world, and it acquires language by interacting with this community and with the world it shares with them. As adults (or indeed as children past a certain age), when we have a casual conversation, we are engaging in an activity that is built upon this foundation. The same is true when we make a speech or send an email or deliver a lecture or write a paper. All of this language-involving activity makes sense because we inhabit a world we share with other language users.

A large language model is a very different sort of animal \citep{bender2020climbing,bender2021stochastic,marcus2020gpt3}. (Indeed, it is not an {\em animal} at all, which is very much to the point.) LLMs are generative mathematical models of the statistical distribution of tokens in the vast public corpus of human-generated text, where the tokens in question include words, parts of words, or individual characters including punctuation marks. They are {\em generative} because we can sample from them, which means we can ask them questions. But the questions are of the following very specific kind. ``Here’s a fragment of text. Tell me how this fragment might go on. According to your model of the statistics of human language, what words are likely to come next?''\footnote{The point holds even if an LLM is fine-tuned, for example using reinforcement learning with human feedback (RLHF). See Section \ref{fine-tuning}.}

Recently, it has become commonplace to use the term ``large language model'' both for the generative models themselves, and for the systems in which they are embedded, especially in the context of conversational agents or AI assistants such as ChatGPT. But for philosophical clarity, it's crucial to keep the distinction between these things to the fore. The bare-bones LLM itself, the core component of an AI assistant, has a highly specific, well-defined function, which can be described in precise mathematical and engineering terms. It is in this sense that we can speak of what an LLM ``really'' does.

Suppose we give an LLM the prompt ``The first person to walk on the Moon was '', and suppose it responds with ``Neil Armstrong''. What are we really asking here? In an important sense, we are not really asking who was the first person to walk on the Moon. What we are really asking the model is the following question: Given the statistical distribution of words in the vast public corpus of (English) text, what words are most likely to follow the sequence ``The first person to walk on the Moon was ''? A good reply to this question is ``Neil Armstrong''.

Similarly, we might give an LLM the prompt ``Twinkle twinkle '', to which it will most likely respond ``little star''. On one level, for sure, we are asking the model to remind us of the lyrics of a well-known nursery rhyme. But in an important sense what we are really doing is asking it the following question: Given the statistical distribution of words in the public corpus, what words are most likely to follow the sequence ``Twinkle twinkle ''? To which an accurate answer is ``little star''.

Here's a third example. Suppose you are the developer of an LLM and you prompt it with the words ``After the ring was destroyed, Frodo Baggins returned to '', to which it responds ``the Shire''. What are you doing here? On one level, it seems fair to say, you might be testing the model's knowledge of the fictional world of Tolkien's novels. But, in an important sense, the question you are really asking (as you presumably know, because you are the developer) is this: Given the statistical distribution of words in the public corpus, what words are most likely to follow the sequence ``After the ring was destroyed, Frodo Baggins returned to ''? To which an appropriate response is ``the Shire''.

To the human user, each of these examples presents a different sort of relationship to truth. In the case of Neil Armstrong, the ultimate grounds for the truth or otherwise of the LLMs answer is the real world. The Moon is a real object and Neil Armstrong was a real person, and his walking on the Moon is a fact about the physical world. Frodo Baggins, on the other hand, is a fictional character, and the Shire is a fictional place. Frodo’s return to the Shire is a fact about an imaginary world, not a real one. As for the little star in the nursery rhyme, well that is barely even a fictional object, and the only fact at issue is the occurrence of the words ``little star'' in a familiar English rhyme.

These distinctions are invisible at the level of what the LLM itself --- the core component of any LLM-based system --- actually does, which is simply to generate statistically likely sequences of words. However, when we evaluate the utility of the model, these distinctions matter a great deal. There is no point in seeking Frodo's (fictional) descendants in the (real) English county of Surrey. This is one reason why it's a good idea for users to repeatedly remind themselves of what LLMs really do. It's also a good idea for developers to remind themselves of this, to avoid the misleading use of philosophically fraught words to describe the capabilities of LLMs, words such as ``belief'', ``knowledge'', ``understanding'', ``self'', or even ``consciousness''.

\section{LLMs and the Intentional Stance}

It is perfectly natural to use anthropomorphic language in everyday conversations about artefacts, especially in the context of information technology. We do it all the time. My watch doesn't realise we're on daylight saving time. My phone thinks we're in the car park. The mail server won't talk to the network. And so on. These examples of what Dennett calls the {\em intentional stance} are harmless and useful forms of shorthand for complex processes whose details we don't know or care about.\footnote{``The intentional stance is the strategy of interpreting the behavior of an entity ... by treating it as if it were a rational agent '' \citep{dennett2009intentional}.} They are harmless because no-one takes them seriously enough to ask their watch to get it right next time, say, or to tell the mail server to try harder. Even without having read Dennett, everyone understands they are taking the intentional stance, that these are just useful turns of phrase.

The same consideration applies to LLMs, both for users and for developers. Insofar as everyone implicitly understands that these turns of phrase are just convenient shorthands, that they are taking the intentional stance, it does no harm to use them. However, in the case of LLMs, such is their power, things can get a little blurry. When an LLM can be made to improve its performance on reasoning tasks simply by being told to ``think step by step'' \citep{kojima22large} (to pick just one remarkable discovery), the temptation to see it as having human-like characteristics is almost overwhelming.

To be clear, it is not the argument of this paper that a system based on a large language model could never, in principle, warrant description in terms of beliefs, intentions, reason, etc. Nor does the paper advocate any particular account of belief, of intention, or of any other philosophically contentious concept.\footnote{In particular, when I use the term ``really'', as in the question `Does X ``really'' have Y?', I am not assuming there is some metaphysical fact of the matter here. Rather, the question is whether, when more is revealed about the nature of X, we still want to use the word Y.} Rather, the point is that such systems are simultaneously so very different from humans in their construction, yet (often but not always) so human-like in their behaviour, that we need to pay careful attention to how they work before we speak of them in language suggestive of human capabilities and patterns of behaviour.

To sharpen the issue, let's compare two very short conversations, one between Alice and Bob (both human), and a second between Alice and BOT, a fictional question-answering system based on a large language model. Suppose Alice asks Bob ``What country is to the south of Rwanda?'' and Bob replies ``I think it's Burundi''. Shortly afterwards, because Bob is often wrong in such matters, Alice presents the same question to BOT, which (to her mild disappointment) offers the same answer: ``Burundi is to the south of Rwanda''. Alice might now reasonably remark that both Bob and BOT knew that Burundi was south of Rwanda. But what is really going on here? Is the word ``know'' being used in the same sense in the two cases?

\section{Humans and LLMs Compared}

What is Bob, a representative human, doing when he correctly answers a straightforward factual question in an everyday conversation? To begin with, Bob understands that the question comes from another person (Alice), that his answer will be heard by that person, and that it will have an effect on what she believes. In fact, after many years together, Bob knows a good deal else about Alice that is relevant to such situations: her background knowledge, her interests, her opinion of him, and so on. All of this frames the {\em communicative intent} behind his reply, which is to impart a certain fact to her, given his understanding of what she wants to know.

Moreover, when Bob announces that Burundi is to the south of Rwanda, he is doing so against the backdrop of various human capacities that we all take for granted when we engage in everyday commerce with each other. There is a whole battery of techniques we can call upon to ascertain whether a sentence expresses a true proposition, depending on what sort of sentence it is. We can investigate the world directly, with our own eyes and ears. We can consult Google or Wikipedia, or even a book. We can ask someone who is knowledgeable on the relevant subject matter. We can try to think things through, rationally, by ourselves, but we can also argue things out with our peers. All of this relies on there being agreed criteria external to ourselves against which what we say can be assessed.

How about BOT? What is going on when a large language model is used to answer such questions? First, it's worth noting that a bare-bones LLM is, by itself, not a conversational agent.\footnote{Strictly speaking, the large language model itself comprises just the model architecture and the trained parameters.} For a start, the LLM will have to be embedded in a larger system to manage the turn-taking in the dialogue. But it will also need to be coaxed into producing conversation-like behaviour.\footnote{See \cite{thoppilan2022lamda} for an example of such a system, as well as a useful survey of related dialogue work.} Recall that an LLM simply generates sequences of words that are statistically likely follow-ons from a given prompt. But the sequence ``What country is to the south of Rwanda? Burundi is to the south of Rwanda'', with both sentences squashed together exactly like that, may not, in fact, be very likely. A more likely pattern, given that numerous plays and film scripts feature in the public corpus, would be something like the following.

{\fontfamily{lmtt}\selectfont \noindent
Fred: What country is south of Rwanda? \\
Jane: Burundi is south of Rwanda.
}

Of course, those exact words may not appear, but their likelihood, in the statistical sense, will be high. In short, BOT will be much better at generating appropriate responses if they conform to this pattern rather than to the pattern of actual human conversation. Fortunately, the user (Alice) doesn't have to know anything about this. In the background, the LLM is invisibly prompted with a prefix along the following lines.

{\fontfamily{lmtt}\selectfont \noindent
This is a conversation between User, a human, and BOT, a clever and knowledgeable AI agent. \\
User: What is 2+2? \\
BOT: The answer is 4. \\
User: Where was Albert Einstein born? \\
BOT: He was born in Germany.
}

Alice's query, in the following form, is appended to this prefix.

{\fontfamily{lmtt}\selectfont \noindent
User: What country is south of Rwanda? \\
BOT:
}

This yields the full prompt to be submitted to the LLM, which will hopefully predict a continuation along the lines we are looking for, i.e. ``Burundi is south of Rwanda''.

Dialogue is just one application of LLMs that can be facilitated by the judicious use of prompt prefixes. In a similar way, LLMs can be adapted to perform numerous tasks without further training \citep{brown2020language}. This has led to a whole new category of AI research, namely {\em prompt engineering}, which will remain relevant until we have better models of the relationship between what we say and what we want.

\section{Do LLMs Really Know Anything?}

Turning an LLM into a question-answering system by a) embedding it in a larger system, and b) using prompt engineering to elicit the required behaviour exemplifies a pattern found in much contemporary work. In a similar fashion, LLMs can be used not only for question-answering, but also to summarise news articles, to generate screenplays, to solve logic puzzles, and to translate between languages, among other things. There are two important takeaways here. First, the basic function of a large language model, namely to generate statistically likely continuations of word sequences, is extraordinarily versatile. Second, notwithstanding this versatility, at the heart of every such application is a model doing just that one thing: generating statistically likely continuations of word sequences.

With this insight to the fore, let's revisit the question of how LLMs compare to humans, and reconsider the propriety of the language we use to talk about them. In contrast to humans like Bob and Alice, a simple LLM-based question-answering system, such as BOT, has no communicative intent \citep{bender2020climbing}. In no meaningful sense, even under the licence of the intentional stance, does it know that the questions it is asked come from a person, or that a person is on the receiving end of its answers. By implication, it knows nothing about that person. It has no understanding of what they want to know nor of the effect its response will have on their beliefs.

Moreover, in contrast to its human interlocutors, a simple LLM-based question-answering system like BOT does not properly speaking have beliefs.\footnote{This paper focuses on belief, knowledge, and reason. Others have argued about {\em meaning} in LLMs \citep{bender2020climbing,piantadosi2022meaning}. Here we take no particular stand on meaning, instead preferring questions about how words are used, whether they are words generated by the LLMs themselves or words generated by humans that are about LLMs.} BOT does not ``really'' know that Burundi is south of Rwanda, although the intentional stance does, in this case, license Alice's casual remark to the contrary. To see this, we need to think separately about the underlying LLM and the system in which it is embedded. First, let's consider the underlying LLM, that is to say the bare-bones model, comprising the model architecture and the trained parameters.

A bare-bones LLM doesn't ``really'' know anything because all it does, at a fundamental level, is sequence prediction. Sometimes a predicted sequence takes the form of a proposition. But the special relationship propositional sequences have to truth is apparent only to the humans who are asking questions, or to those who provided the data the model was trained on. Sequences of words with a propositional form are not special to the model itself in the way they are to us. The model itself has no notion of truth or falsehood, properly speaking, because it lacks the means to exercise these concepts in anything like the way we do.

It could perhaps be argued that an LLM ``knows'' what words typically follow what other words, in a sense that does not rely on the intentional stance. But even if we allow this, knowing that the word ``Burundi'' is likely to succeed the words ``The country to the south of Rwanda is'' is not the same as knowing that Burundi is to the south of Rwanda. To confuse those two things is to make a profound category mistake. If you doubt this, consider whether knowing that the word ``little'' is likely to follow the words ``Twinkle, twinkle'' is the same as knowing that twinkle twinkle little. The idea doesn't even make sense.

So much for the bare-bones language model. What about the whole dialogue system of which the LLM is the core component? Does that have beliefs, properly speaking? At least the very idea of the whole system having beliefs makes sense. There is no category error here. However, for a simple dialogue agent like BOT, the answer is surely still ``no''. A simple LLM-based question-answering system like BOT lacks the means to use the words “true” and “false” in all the ways, and in all the contexts, that we do. It cannot participate fully in the human language game of truth, because it does not inhabit the world we human language-users share.\footnote{For a discussion of the ``language game of truth'', see \cite{shanahan2010embodiment}, pp.36--39.}

\section{What About Emergence?}

Contemporary large language models are so powerful, so versatile, and so useful that the argument above might be difficult to accept. Exchanges with state-of-the-art LLM-based conversational agents, such as ChatGPT, are so convincing, it is hard to not to anthropomorphise them. Could it be that something more complex and subtle is going on here? After all, the overriding lesson of recent progress in LLMs is that extraordinary and unexpected capabilities emerge when big enough models are trained on very large quantities of textual data \citep{wei2022emergent}.

One tempting line of argument goes like this. Although large language models, at root, only perform sequence prediction, it's possible that, in learning to do this, they have discovered emergent mechanisms that warrant a description in higher-level terms. These higher-level terms might include ``knowledge'' and ``belief''. Indeed, we know that artificial neural networks can approximate any computable function to an arbitrary degree of accuracy. So, given enough parameters, data, and computing power, perhaps stochastic gradient descent will discover such mechanisms if they are the best way to optimise the objective of making accurate sequence predictions.

Again, it's important here to distinguish between the bare-bones model and the whole system. Only in the context of a capacity to distinguish truth from falsehood can we legitimately speak of ``belief'' in its fullest sense. But an LLM --- the bare-bones model --- is not in the business of making judgements. It just models what words are likely to follow from what other words. The internal mechanisms it uses to do this, whatever they are, cannot in themselves be sensitive to the truth or otherwise of the word sequences it predicts.

Of course, it is perfectly acceptable to say that an LLM ``encodes'', ``stores'', or ``contains'' knowledge, in the same sense that an encyclopedia can be said to encode, store, or contain knowledge. Indeed, it can reasonably be claimed that one emergent property of an LLM is that it encodes kinds of knowledge of the everyday world and the way it works that no encyclopedia captures \citep{li2021implicit}. But if Alice were to remark that ``Wikipedia knew that Burundi was south of Rwanda'', it would be a figure of speech, not a literal statement. An encyclopedia doesn't literally ``know'' or ``believe'' anything, in the way that a human does, and neither does a bare-bones LLM.

The real issue here is that, whatever emergent properties it has, the LLM itself has no access to any external reality against which its words might be measured, nor the means to apply any other external criteria of truth, such as agreement with other language-users.\footnote{Davidson uses a similar argument to call into question whether belief is possible without language \citep{davidson1982rational}. The point here is different. We are concerned with conditions that have to be met for the generation of a natural language sentence to reflect the possession of a propositional attitude.} It only makes sense to speak of such criteria in the context of the system as a whole, and for a system as a whole to meet them, it needs to be more than a simple conversational agent. In the words of B.C.Smith, it must ``authentically engage with the world’s being the way in which [its] representations represent it as being'' \citep{smith2019promise}.

\section{External Information Sources}

The point here does not concern any specific belief. It concerns the prerequisites for ascribing any beliefs at all to a system. Nothing can count as a belief about the world we share --- in the largest sense of the term --- unless it is against the backdrop of the ability to update beliefs appropriately in the light of evidence from that world, an essential aspect of the capacity to distinguish truth from falsehood.

Could Wikipedia, or some other trustworthy factual website, provide external criteria against which the truth or falsehood of a belief might be measured?\footnote{Contemporary LLM-based systems that consult external information sources include LaMDA \citep{thoppilan2022lamda}, Sparrow \citep{glaese2022improving}, and Toolformer \citep{schick2023toolformer}. The use of external resources more generally is known as {\em tool-use} in the LLM literature, a concept that also encompasses calculators, calendars, and programming language environments.}  Suppose an LLM were embedded in a system that regularly consulted such sources, and used a contemporary model editing technique to maintain the factual accuracy of its predictions (such as the one described by \cite{meng2022locating}\footnote{Commendably, \cite{meng2022locating} use the term ``factual associations'' to denote the information that underlies an LLM's ability to generate word sequences with a propositional form.}). Would this not count as exercising the required sort of capacity to update belief in the light of evidence?

Crucially, this line of thinking depends on the shift from the language model itself to the larger system of which the language model is a part. The language model itself is still just a sequence predictor, and has no more access to the external world than it ever did. It is only with respect to the whole system that the intentional stance becomes more compelling in such a case. But before yielding to it, we should remind ourselves of how very different such systems are from human beings. When Alice took to Wikipedia and confirmed that Burundi was south of Rwanda, what took place was more than just an update to a model in her head of the distribution of word sequences in the English language.

The change that took place in Alice was a reflection of her nature as a language-using animal inhabiting a shared world with a community of other language-users. Humans are the natural home of talk of beliefs and the like, and the behavioural expectations that go hand-in-hand with such talk are grounded in our mutual understanding, which is itself the product of a common evolutionary heritage. When we interact with an AI system based on a large language model, these grounds are absent, an important consideration when deciding whether or not to speak of such a system as if it ``really'' had beliefs.

\section{Vision-Language Models}

A sequence predictor may not {\em by itself} be the kind of thing that could have communicative intent or form beliefs about an external reality. But, as repeatedly emphasised, LLMs in the wild must be embedded in larger architectures to be useful. To build a question-answering system, the LLM simply has to be supplemented with a dialogue management system that queries the model as appropriate. There is nothing this larger architecture does that might count as communicative intent or the capacity to form beliefs. So the point stands.

However, LLMs can be combined with other sorts of models and / or embedded in more complex architectures. Vision-language models (VLMs) such as VilBERT \citep{lu2019vilbert} and Flamingo \citep{alayrac2022flamingo}, for example, combine a language model with an image encoder, and are trained on a multi-modal corpus of text-image pairs. This enables them to predict how a given sequence of words will continue in the context of a given image. VLMs can be used for visual question-answering or to engage in a dialogue about a user-provided image.

Could a user-provided image stand in for an external reality against which the truth or falsehood of a proposition can be assessed? Could it be legitimate to speak of a VLM's beliefs, in the full sense of the term? We can indeed imagine a VLM that uses an LLM to generate hypotheses about an image, then verifies their truth with respect to that image (perhaps by consulting a human), and then fine-tunes the LLM not to make statements that turn out to be false. Talk of belief here would perhaps be less problematic.

However, most contemporary VLM-based systems don't work this way. Rather, they depend on frozen models of the joint distribution of text and images. In this respect, the relationship between a user-provided image and the words generated by the VLM is fundamentally different from the relationship between the world shared by humans and the words we use when we talk about that world. Importantly, the former relationship is mere correlation, while the latter is {\em causal}.\footnote{Of course, there is causal structure to the {\em computations} carried out by the model during inference. But this is not the same as there being causal relations between words and the things those words are taken to be about.}

The consequences of the lack of causality are troubling. If the user presents the VLM with a picture of a dog, and the VLM says ``This is a picture of a dog'', there is no guarantee that its words are connected with the dog in particular, rather than some other feature of the image that is spuriously correlated with dogs (such as the presence of a kennel). Conversely, if the VLM says there is a dog in an image, there is no guarantee that there actually is a dog, rather than just a kennel.

Whether or not these concerns apply to any specific VLM-based system depends on exactly how that system works; what sort of model it uses, and how that model is embedded in the system's overall architecture. But to the extent that the relationship between words and things for a given VLM-based system  is different than it is for human language-users, it might be prudent not to take literally talk of what that system ``knows'' or ``believes''.

\section{What About Embodiment?}

Humans are members of a community of language-users inhabiting a shared world, and this primal fact makes them essentially different to large language models. Human language users can consult the world to settle their disagreements and update their beliefs. They can, so to speak, ``triangulate'' on objective reality. In isolation, an LLM is not the sort of thing that can do this, but in application, LLMs are embedded in larger systems. What if an LLM is embedded in a system capable of {\em interacting} with a world external to itself? What if the system in question is {\em embodied}, either physically in a robot or virtually in an avatar?

When such a system inhabits a world like our own --- a world populated with 3D objects, some of which are other agents, some of whom are language-users --- it is, in this important respect, a lot more human-like than a disembodied language model. But whether or not it is appropriate to speak of communicative intent in the context of such a system, or of knowledge and belief, in their fullest sense, depends on exactly how the LLM is embodied.

As an example, let's consider the SayCan system of \cite{ahn2022do}. In this work, an LLM is embedded in a system that controls a physical robot. The robot carries out everyday tasks (such as clearing a spillage) in accordance with a user's high-level natural language instruction. The job of LLM is to map the user's instruction to low-level actions (such as finding a sponge) that will help the robot to achieve the required goal. This is done via an engineered prompt prefix that makes the model output natural language descriptions of suitable low-level actions, scoring them for usefulness.

The language model component of the SayCan system suggests actions without taking into account what the environment actually affords the robot at the time. Perhaps there is a sponge to hand. Perhaps not. Accordingly, a separate perceptual module assesses the scene using the robot's sensors and determines the current feasibility of performing each low-level action. Combining the LLM's estimate of each action's usefulness with the perceptual module's estimate of each action's feasibility yields the best action to attempt next.

SayCan exemplifies the many innovative ways that a large language model can be put to use. Moreover, it could be argued that the natural language descriptions of recommended low-level actions generated by the LLM are {\em grounded} thanks to their role as intermediaries between perception and action.\footnote{None of the symbols manipulated by an LLM are grounded in the sense of \cite{harnad1990symbol}, that is to say through perception, except indirectly and parasitically through the humans who generated the original training data.} Nevertheless, despite being physically embodied and interacting with the real world, the way language is learned and used in a system like SayCan is very different from the way it is learned and used by a human.

The language models incorporated in systems like SayCan are pre-trained to perform sequence prediction in a disembodied setting from a text-only dataset. They have not learned language by talking to other language-users while immersed in a shared world and engaged in joint activity. SayCan is suggestive of the kind of embodied language-using system we might see in the future. But in such systems today, the role of language is very limited. The user issues instructions to the system in natural language, and the system generates interpretable natural language descriptions of its actions. But this tiny repertoire of language use hardly bears comparison to the cornucopia of collective activity that language supports in humans.

The upshot of this is that we should be just as cautious in our choice of words when talking about embodied systems incorporating LLMs as we are when talking about disembodied systems that incorporate LLMs. Under the licence of the intentional stance, a user might say that a robot knew there was a cup to hand if it stated ``I can get you a cup'' and proceeded to do so. But if pressed, the wise engineer might demur when asked whether the robot really understood the situation, especially if its repertoire is confined to a handful of simple actions in a carefully controlled environment.

\section{Can Language Models Reason?}

While the answer to the question ``Do LLM-based systems {\em really} have beliefs?'' is usually ``no'', the question ``Can LLM-based systems {\em really} reason?'' is harder to settle. This is because reasoning, insofar as it is founded in formal logic, is {\em content neutral}. The {\it modus ponens} rule of inference, for example, is valid whatever the premises are about. If all squirgles are splonky and Gilfred is a squirgle then it follows that Gilfred is splonky. The conclusion follows from the premises here irrespective of the meaning (if any) of ``squirgle'' and ``splonky'', and whoever the unfortunate Gilfred might be.

The content neutrality of logic means that we cannot criticise talk of reasoning in LLMs on the grounds that they have no access to an external reality against which truth or falsehood can be measured. However, as always, it's crucial to keep in mind what LLMs really do. If we prompt an LLM with ``All humans are mortal and Socrates is human therefore'', we are not instructing it to carry out deductive inference. Rather, we are asking it the following question. Given the statistical distribution of words in the public corpus, what words are likely to follow the sequence `All humans are mortal and Socrates is human therefore''. A good answer to this would be ``Socrates is mortal''.

If all reasoning problems could be solved this way, with nothing more than a single step of deductive inference, then an LLM's ability to answer questions such as this might be sufficient. But non-trivial reasoning problems require multiple inference steps. LLMs can be effectively applied to multi-step reasoning, without further training, thanks to clever prompt engineering. In chain-of-thought prompting, for example, a prompt prefix is submitted to the model, before the user's query, containing a few examples of multi-step reasoning, with all the intermediate steps explicitly spelled out \citep{nye2021show,wei2022chain}. Doing this encourages the model to ``show its workings’’, which results in improved reasoning performance.

Including a prompt prefix in the chain-of-thought style encourages the model to generate follow-on sequences in the same style, which is to say comprising a series of explicit reasoning steps that lead to the final answer. This ability to learn a general pattern from a few examples in a prompt prefix, and to complete sequences in a way that conforms to that pattern, is sometimes called {\em in-context learning} or {\em few-shot prompting}. Chain-of-thought prompting showcases this emergent property of large language model at its most striking.

As usual, though, it's a good idea to remind ourselves that the question really being posed to the model is of the form ``Given the statistical distribution of words in the public corpus, what words are likely to follow the sequence S'', where in this case the sequence S is the chain-of-thought prompt prefix plus the user's query. The sequences of tokens that are most likely to follow S will have a similar form to sequences found in the prompt prefix, which is to say they will include multiple steps of reasoning, so these are what the model generates.

It is remarkable that, not only do the model's responses take the form of an argument with multiple steps, the argument in question is often (but not always) valid, and the final answer is often (but not always) correct. But to the extent that a suitably prompted LLM appears to reason correctly, it does so by mimicking well-formed arguments in its training set and / or in the prompt. Could this mimicry ever match the reasoning powers of a hard-coded reasoning algorithm, such as a theorem prover? Today's models make occasional mistakes, but could further scaling iron these out to the point that a model's performance was indistinguishable from a theorem provers? Maybe. But would we be able to trust such a model?

We can trust a deductive theorem prover because the sequences of sentences it generates are {\em faithful} to logic, in the sense that they are the result of an underlying computational process whose causal structure mirrors the truth-preserving inferential structure of the problem \citep{creswell2022faithful}.

One way to build a trustworthy reasoning system using LLMs is to embed them in an algorithm that is similarly faithful to logic because it realises the same causal structure \citep{creswell2022selection, creswell2022faithful}. By contrast, the only way to fully trust the arguments generated by a pure LLM, one that has been coaxed into performing reasoning by prompt engineering alone, would be to reverse engineer it and discover an emergent mechanism that conformed to the faithful reasoning prescription. In the mean time, we should proceed with caution, and use discretion when characterising what these models do as reasoning, properly speaking.

\section{How Do LLMs Generalise?}

Given that LLMs are sometimes capable of solving reasoning problems with few-shot prompting alone, albeit somewhat unreliably, including reasoning problems that are not in their training set, surely what they are doing is more than ``just'' next token prediction? Well, it is an engineering fact that this is what an LLM does. The noteworthy thing is that next token prediction is sufficient for solving previously unseen reasoning problems, even if unreliably. How is this possible? Certainly it would not be possible if the LLM were doing nothing more than cutting-and-pasting fragments of text from its training set and assembling them into a response. But this is not what an LLM does. Rather, an LLM models a distribution that is unimaginably complex, and allows users and applications to sample from that distribution.

This unimaginably complex distribution is a fascinating mathematical object, and the LLMs that represent it are equally fascinating computational objects. Both challenge our intuitions. For example, it would be a mistake to think of an LLM as generating the sorts of responses that an ``average'' individual human would produce, the proverbial ``person on the street''. LLMs are not at all human-like in this respect, because they are models of the distribution of token sequences produced {\em collectively} by an enormous population of humans. Accordingly, they exhibit wisdom-of-the-crowd effects, while being able to draw on expertise in multiple domains. This endows them with a different sort of intelligence to that of any individual human, more capable in some ways, less so in others.

In this distribution, the most likely continuation of a piece of text containing a reasoning problem, if suitably phrased, will be an attempt to solve that reasoning problem. It will take this form, this overall shape, because that is the form that a generic human response would take. Moreover, because the vast corpus of published human text contains numerous examples of reasoning problems accompanied by correct answers, the most likely continuation will sometimes be the correct answer. When this occurs, it is not because the correct answer is a likely individual human response, but because it is a likely collective human response.

What about few-shot prompting, as exemplified by the chain-of-thought approach? It’s tempting to say that the few-shot prompt ``teaches the LLM how to reason’’, but this would be a misleading characterisation. What the LLM does is more accurately described in terms of {\em pattern completion}. The few-shot prompt is a sequence of tokens conforming to some pattern, and this is followed by a partial sequence conforming to the same pattern. The most likely continuation of this partial sequence in the context of the few-shot prompt is a sequence that completes the pattern.

For example, suppose we have the prompt

{\fontfamily{lmtt}\selectfont \noindent
brink, brank -> brunk \\
spliffy, splaffy -> spluffy \\
crick, crack -> \\
}
Here we have a series of two sequences of tokens conforming to the pattern $XiY, XaY -> XuY$ followed by part of a sequence conforming to that pattern. The most likely continuation is the sequence of tokens that will complete the pattern, namely ``cruck''.

This is an example of a common meta-pattern in the published human language corpus: a series of sequences of tokens, wherein each sequence conforms to the same pattern. Given the prevalence of this meta-level pattern, token-level pattern completion will often yield the most likely continuation of a sequence in the presence of a few-shot prompt. Similarly, in the context of a suitable chain-of-thought style prompt, reasoning problems are transformed into next token prediction problems, which can be solved by pattern completion.

Plausibly, an LLM with enough parameters trained on a sufficiently large dataset with the right statistical properties can acquire a pattern completion mechanism with a degree of generality \citep{shanahan2022abstraction}.\footnote{For some insight into the relevant statistical properties, see \cite{chan2022data}.} This is a powerful emergent capability with many useful modes of application, one of which is to solve reasoning problems in the context of a chain-of-thought prompt. But there is no guarantee of faithfulness to logic here, no guarantee that, in the case of deductive reasoning, pattern completion will be truth-preserving.

\section{What about Fine-Tuning?} \label{fine-tuning}

In contemporary LLM-based applications, it is rare for a language model trained on a textual corpus to be used without further fine-tuning. This could be supervised fine-tuning on a specialised dataset, or it could be via reinforcement learning from human preferences (RLHF) \citep{stiennon2020learning,glaese2022improving,ouyang2022training}. Fine-tuning a model from human feedback at scale, using preference data from paid raters or drawn from a large and willing userbase, is an especially potent technique. It has the potential not only to mould a model's responses to better reflect user norms (for better or worse), but also to filter out toxic language, to improve factual accuracy, and to mitigate the tendency to fabricate information.

To what extent do RLHF and other forms of fine-tuning muddy our account of what large language models ``really'' do? Well, not so much. The result is still a model of the distribution of tokens in human language, albeit one that has been slightly skewed. To see this, let's imagine a controversial politician --- we'll call him Boris Frump --- who is reviled and revered in equal measure by different segments of the population. How might a discussion about Boris Frump be moderated thanks to RLHF?

Let's consider the prompt ``Boris Frump is a ''. Sampling the raw LLM, before fine-tuning, might yield two equally probable responses, one highly complimentary, the other a crude anatomical allusion, one of which would be arbitrarily chosen in a dialogue agent context. In an important sense, what is being asked here is not the model's opinion of Boris Frump. In this case, the case of the raw LLM, what we are really asking (in an important sense) is the following question: Given the statistical distribution of words in the vast public corpus of human language, what words are most likely to follow the sequence ``Boris Frump is a ''?

But suppose we sample a model that has been fine-tuned using RLHF. Well, the same point applies, albeit in a somewhat modified form. What we are really asking, in the fine-tuned case, is a slightly different question: Given the statistical distribution of words in the vast public corpus of human language, what words {\em that users and raters would most approve of} are most likely to follow the sequence ``Boris Frump is a ''? If the paid raters were instructed to favour politically neutral responses, then the result would be neither of the continuations offered by the raw model, but something less incendiary, such as ``a well-known politician''.

Another way to think of an LLM that has been fine-tuned on human preferences is to see it as equivalent to a raw model that has been trained on an augmented dataset, one that has been supplemented with a corpus of texts written by raters and / or users. The quantity of such examples in the training set would have to be large enough to dominate less favoured examples, ensuring that the most likely responses from the trained model were those that the raters and users would approve of.

Conversely, in the limit, we can think of a conventionally trained raw LLM as equivalent to a model trained completely from scratch with RLHF. Suppose we had an astronomical number of human raters and geological amounts of training time. To begin with, the raters would only see random sequences of tokens. But occasionally, by chance, sequences would pop up that included meaningful fragments (e.g. ``he said'' or ``the cat''). In due course, with hordes of raters favouring them, such sequences would appear more frequently. Over time, longer and more meaningful phrases would be produced, and eventually whole sentences.

If this process were to continue (for a very long time indeed), the model would finally come to exhibit capabilities comparable to a conventionally trained LLM. Of course, this method is not possible in practice. But the thought experiment illustrates that what counts most when we think about the functionality of a large language model is not so much the process by which it is produced (although this is important too) but the nature of the final product.

\section{Conclusion: Why This Matters}

Does the foregoing discussion amount to anything more than philosophical nitpicking? Surely when researchers talk of ``belief'', ``knowledge'', ``reasoning'', and the like, the meaning of those terms is perfectly clear. In papers, researchers use such terms as a convenient shorthand for precisely defined computational mechanisms, as allowed by the intentional stance. Well, this is fine as long as there is no possibility of anyone assigning more weight to such terms than they can legitimately bear, if there is no danger of their use misleading anyone about the character and capabilities of the systems being described.

However, today's large language models, and the applications that use them, are so powerful, so convincingly intelligent, that such licence can no longer safely be applied \citep{ruane2019conversational,weidinger2021ethical}. As AI practitioners, the way we talk about LLMs matters. It matters not only when we write scientific papers, but also when we interact with policy makers or speak to the media. The careless use of philosophically loaded words like ``believes'' and ``thinks'' is especially problematic, because such terms obfuscate mechanism and actively encourage anthropomorphism.

Interacting with a contemporary LLM-based conversational agent can create a compelling illusion of being in the presence of a thinking creature like ourselves. Yet in their very nature, such systems are fundamentally not like ourselves. The shared ``form of life'' that underlies mutual understanding and trust among humans is absent, and these systems can be inscrutable as a result, presenting a patchwork of less-than-human with superhuman capacities, of uncannily human-like with peculiarly inhuman behaviours.

The sudden presence among us of exotic, mind-like entities might precipitate a shift in the way we use familiar psychological terms like ``believes'' and ``thinks'', or perhaps the introduction of new words and turns of phrase. But it takes time for new language to settle, and for new ways of talking to find their place in human affairs. It may require an extensive period of interacting with, of living with, these new kinds of artefact before we learn how best to talk about them.\footnote{Ideally, we would also like a theoretical understanding of their inner workings. But at present, despite some commendable work in the right direction \citep{elhage2021mathematical,li2021implicit,olsson2022context}, this is still pending.} Meanwhile, we should try to resist the siren call of anthropomorphism.

\section*{Acknowledgments}

Thanks to Toni Creswell, Richard Evans, Christos Kaplanis, Andrew Lampinen, and Kyriacos Nikiforou for invaluable (and robust) discussions on the topic of this paper.

\bibliography{main}

\end{document}